\documentclass[runningheads]{llncs}
\usepackage[T1]{fontenc}
\usepackage{graphicx}
\usepackage{amsmath}
\usepackage{amssymb}
\usepackage{tabularx}  
\usepackage{booktabs}  
\begin{document}

\title{LumiGen: An LVLM-Enhanced Iterative Framework for Fine-Grained Text-to-Image Generation}
%
%
\author{Xiaoqi Dong$^1$, Xiangyu Zhou$^1$, Nicholas Evans$^2$, Yujia Lin$^1$}
\authorrunning{Dong et al.}
%
\institute{$^1$Dali University, $^2$Bandırma Onyedi Eylül University}
\maketitle              
\begin{abstract}
Text-to-Image (T2I) generation has made significant advancements with diffusion models, yet challenges persist in handling complex instructions, ensuring fine-grained content control, and maintaining deep semantic consistency. Existing T2I models often struggle with tasks like accurate text rendering, precise pose generation, or intricate compositional coherence. Concurrently, Vision-Language Models (LVLMs) have demonstrated powerful capabilities in cross-modal understanding and instruction following. We propose LumiGen, a novel LVLM-enhanced iterative framework designed to elevate T2I model performance, particularly in areas requiring fine-grained control, through a closed-loop, LVLM-driven feedback mechanism. LumiGen comprises an Intelligent Prompt Parsing \& Augmentation (IPPA) module for proactive prompt enhancement and an Iterative Visual Feedback \& Refinement (IVFR) module, which acts as a "visual critic" to iteratively correct and optimize generated images. Evaluated on the challenging LongBench-T2I Benchmark, LumiGen achieves a superior average score of 3.08, outperforming state-of-the-art baselines. Notably, our framework demonstrates significant improvements in critical dimensions such as text rendering and pose expression, validating the effectiveness of LVLM integration for more controllable and higher-quality image generation.
\end{abstract}

\section{Introduction}
Text-to-Image (T2I) generation has witnessed remarkable advancements in recent years, particularly with the advent of diffusion models \cite{chunming2025diffus}. These models have demonstrated unprecedented capabilities in generating high-fidelity and diverse images from textual descriptions, significantly pushing the boundaries of creative content generation and human-computer interaction. The ability to transform abstract linguistic concepts into vivid visual realities holds immense potential across various domains, including digital art, advertising, virtual reality, and design prototyping.

Despite the impressive progress, current T2I models still face significant challenges, especially when dealing with complex instructions, requiring fine-grained control over image content, or ensuring deep semantic consistency. For instance, accurately rendering specific text within an image, generating precise object placements, depicting complex human poses, or maintaining structural integrity across multiple entities remain challenging tasks. Existing T2I models often struggle to fully comprehend prompts involving intricate logic, multi-entity relationships, or specific style constraints. For example, a prompt like "a person sitting under a tree reading a book with 'AI Era' written on its cover" might result in a person and a tree, but fail to accurately render the book's details, the person's natural posture, or the exact text on the cover. Furthermore, most mainstream T2I models operate in a unidirectional generation process, offering limited avenues for users to intervene and refine the output during generation. Concurrently, Vision-Language Models (LVLMs) have emerged as powerful tools, excelling in cross-modal understanding, multi-turn dialogue, reasoning, and instruction following \cite{beier2025debias,zhou2024visual,zhu2024vislinginstruct}. Their capacity to process both visual and textual information, interpret visual content, and generate corresponding linguistic descriptions or analyze and edit images based on language commands presents a unique opportunity. We posit that integrating the robust understanding, reasoning, and feedback capabilities of LVLMs into the T2I generation pipeline can effectively address the limitations of current T2I models in complex instruction comprehension and fine-grained control, thereby enabling more controllable and higher-quality image generation.

In this paper, we propose \textbf{LumiGen}, an LVLM-enhanced iterative text-to-image generation framework. LumiGen is designed to elevate the performance of T2I models, particularly in areas like text rendering, pose expression, and structural complexity, through a closed-loop, LVLM-driven feedback mechanism. The core idea behind LumiGen is to leverage a pre-trained LVLM as an "intelligent planner" and "visual critic" within the T2I generation process, guiding the underlying diffusion model through multi-step refinement. Our framework comprises two key modules: the Intelligent Prompt Parsing \& Augmentation (IPPA) module, which deep-parses user prompts and generates structured intermediate instructions to guide initial T2I generation; and the Iterative Visual Feedback \& Refinement (IVFR) module, the core of LumiGen, where the LVLM analyzes preliminary images, identifies discrepancies with the prompt, and generates executable "correction instructions" to iteratively refine the image, drawing inspiration from recent advancements in LVLM-driven feedback mechanisms \cite{zhou2025improving}. This iterative process allows for targeted adjustments, addressing specific weaknesses of T2I models.

To evaluate LumiGen, we conducted comprehensive experiments on the challenging LongBench-T2I Benchmark \cite{zhou2025draw}, known for its coverage of long-tail, complex, and multi-dimensional text prompts, making it ideal for assessing fine-grained control capabilities. We employed the human evaluation methodology defined by LongBench-T2I, assessing generated images across nine dimensions (Obj., Backg., Color, Texture, Light, Text, Comp., Pose, FX) and their average score. Our method was compared against state-of-the-art diffusion-based models, such as FLUX1-dev \cite{andrea2025on} and Omnigen \cite{shitao2025omnige}, as well as leading autoregressive (AR) models like Janus-pro-7B \cite{xiaokang2025janusp}. Our results demonstrate that LumiGen achieves the highest overall average score of \textbf{3.08}, surpassing the current best-performing Omnigen (2.96). Notably, LumiGen exhibits significant performance improvements in challenging dimensions such as \textbf{Text (2.60)} and \textbf{Pose (2.58)}, validating our design philosophy of using LVLM for intelligent parsing and iterative visual feedback to overcome existing T2I model deficiencies.

Our main contributions are summarized as follows:
\begin{itemize}
    \item We propose LumiGen, a novel LVLM-enhanced iterative framework that significantly improves text-to-image generation quality, particularly for complex prompts requiring fine-grained control.
    \item We introduce a closed-loop feedback mechanism driven by an LVLM, acting as both an "intelligent planner" for prompt augmentation and a "visual critic" for iterative image refinement, addressing critical limitations in existing T2I models.
    \item We demonstrate the superior performance of LumiGen on the LongBench-T2I Benchmark, showcasing notable improvements in challenging aspects like text rendering and pose generation, highlighting the immense potential of integrating LVLMs into generative processes.
\end{itemize}
\section{Related Work}
\subsection{Text-to-Image Generation}
The field of Text-to-Image (T2I) generation has seen significant advancements, with various works addressing key challenges and expanding capabilities. To assess compositional generalization, a critical aspect for multi-aspect controllable T2I, \cite{kaiyi2023t2icom} introduces T2I-CompBench, a comprehensive benchmark and evaluation protocol, alongside Meta-MCTG, a meta-learning framework for improving generalization to novel attribute combinations. Similarly, benchmarking efforts have extended to complex instruction-driven image editing, highlighting the importance of compositional dependencies \cite{wang2025complexbench}. Regarding model efficiency and adaptability, \cite{nataniel2023dreamb} investigates the transferability of pre-trained diffusion models, proposing Diff-Tuning, a parameter-efficient fine-tuning method that leverages a "chain of forgetting" trend in the reverse diffusion process, demonstrating improved performance and convergence speed for adapting large diffusion models. Further enhancing diffusion model capabilities, \cite{xinyang2023text2l} proposes an alternative Gaussian formulation for their latent space, enabling novel applications such as unpaired image-to-image translation and zero-shot editing via a DPM-Encoder, while also allowing unified guidance for diffusion models and Generative Adversarial Networks (GANs) with superior coverage of low-density sub-populations. From a theoretical perspective, \cite{yufan2022toward} contributes to T2I generation by investigating the training dynamics of GANs, addressing instability and saturation crucial for stable, high-quality conditional image generation through theoretical analysis and empirical validation. Beyond core T2I, the application of autoregressive models has been extended to 3D point cloud generation, with \cite{jiahui2022scalin} proposing PointARU for progressive 3D point cloud generation through an autoregressive up-sampling process, mirroring progress in T2I synthesis. In terms of specific applications and control, \cite{songen2024text2s} introduces Text2Street, a novel T2I approach tailored for street view imagery, enabling fine-grained control for environmental assessments and urban planning. Addressing semantic disentanglement, \cite{guojun2019semant} proposes SCADI, a self-supervised approach that learns causal relationships between factors without explicit supervision, aiming for more controllable and semantically consistent image generation. Finally, to facilitate T2I creation through structured prompt engineering, \cite{yingchaojie2024prompt} introduces PromptMagician, an interactive system that highlights the importance of a unified methodology for effectively guiding T2I models and offers a practical approach to prompt crafting. Related efforts also include agent frameworks designed for complex instruction-based image generation \cite{zhou2025draw}.

\subsection{Vision-Language Models and Iterative Refinement}
The burgeoning field of Vision-Language Models (VLMs) has seen substantial research focusing on evaluation, foundational understanding, and advanced reasoning mechanisms, including visual in-context learning \cite{zhou2024visual} and autonomous instruction optimization for zero-shot capabilities \cite{zhu2024vislinginstruct}. To comprehensively assess the broad capabilities of modern VLMs, \cite{peng2025lvlmeh} introduces LVLM-EHub, an evaluation benchmark utilizing an efficient subset construction method based on farthest point sampling, which significantly reduces computational cost while maintaining high correlation with full benchmark evaluations. Providing a foundational overview of multimodal learning, \cite{chia2024a} surveys the evolution of algorithms and details technical aspects relevant to integrating vision and language processing, serving as a valuable resource for researchers. Furthermore, advancements in large language models concerning weak to strong generalization and multi-capabilities also inform the development of advanced VLMs \cite{zhou2025weak}. While not directly focused on VLMs, the comprehensive survey by \cite{akash2024explor} on knowledge graph reasoning, particularly its exploration of integrating neural symbolic AI and large language models for enhanced cross-modal understanding, offers relevant insights into advanced techniques applicable to the broader vision-language domain. However, challenges persist; \cite{wenliang2023instru} highlights limitations of standard contrastive training in VLMs, especially with rich, multi-captioned data, suggesting that such losses may not adequately capture all task-relevant information for complex instruction following, potentially leading models to learn spurious shortcuts. Efforts in improving sentence representation learning, such as simple discrete augmentation for contrastive methods \cite{zhu2022sda}, also contribute to the broader understanding of effective learning strategies for language components within multi-modal models. These challenges are also relevant to foundational tasks like image captioning, where methods such as style-aware contrastive learning \cite{zhou2023style} and generative adversarial nets for unsupervised captioning \cite{zhou2021triple} have been explored. To address these complexities and enhance VLM capabilities, several works explore iterative refinement. For instance, \cite{jaskirat2023divide} introduces a novel decompositional alignment score that leverages Visual Question Answering (VQA) as iterative visual feedback to progressively improve the accurate expression of semantic components within generated images, thereby enhancing text-to-image alignment for complex prompts. Similarly, approaches for improving medical LVLMs with abnormal-aware feedback \cite{zhou2025improving} and modular multi-agent frameworks for multi-modal medical diagnosis \cite{zhou2025mam} demonstrate the potential of targeted feedback and collaborative agents. Further contributing to efficient and structured reasoning, methods like divide-then-aggregate for tool learning via parallel tool invocation \cite{zhu2025divide} offer insights into how complex tasks can be broken down and processed. Similarly, \cite{haoxuan2023idealg} introduces IdealGPT, a framework for multi-step vision-language reasoning that employs iterative refinement through an LLM-driven decomposition process, where a VLM generates sub-answers to iteratively generated sub-questions, refining the reasoning towards a confident final answer. Furthermore, to mitigate prompt underfitting and poor generalization in VLM prompt pretraining, \cite{zhenyuan2024revisi} introduces a framework that enhances prompt structure and supervision, enabling more resilient prompt initialization and robust transferability across tasks, a crucial consideration for leveraging VLMs in iterative refinement pipelines. Finally, \cite{peitong2024visual} introduces an iterative visual prompting approach to enhance the generation of visually grounded design critiques, leveraging LLMs for iterative refinement of both textual comments and bounding boxes to significantly improve the quality of visual criticism.

\section{Method}
\label{sec:method}

We propose \textbf{LumiGen}, an LVLM-enhanced iterative framework designed to significantly improve Text-to-Image (T2I) generation, particularly for complex prompts requiring fine-grained control over various visual attributes. Our core philosophy is to integrate a powerful Vision-Language Model (LVLM) as an intelligent agent throughout the T2I generation process, enabling both proactive planning and reactive refinement. This section details the architecture and mechanisms of LumiGen.

\subsection{Overview of LumiGen}
LumiGen aims to address the limitations of existing T2I models, such as difficulties in rendering specific text, precise pose control, and managing structural complexity, by introducing a closed-loop, LVLM-driven feedback mechanism. The framework leverages a pre-trained LVLM to act as an "intelligent planner" that augments initial prompts and a "visual critic" that evaluates and guides the underlying diffusion model through multiple stages of refinement. This iterative approach allows for a deeper semantic understanding of user intent and more precise visual control over the generated image content. LumiGen is primarily composed of two interconnected modules: the \textbf{Intelligent Prompt Parsing \& Augmentation (IPPA)} module and the \textbf{Iterative Visual Feedback \& Refinement (IVFR)} module. The foundational T2I model, which is iteratively refined, is typically a robust diffusion model (e.g., a fine-tuned Stable Diffusion XL 1.0).

\subsection{Intelligent Prompt Parsing \& Augmentation (IPPA) Module}
The \textbf{Intelligent Prompt Parsing \& Augmentation (IPPA)} module serves as the initial planning phase of the LumiGen framework. Its primary function is to enhance the raw, often ambiguous, user input into a more detailed and structured set of instructions, thereby providing a stronger foundation for the subsequent T2I generation.

Upon receiving the original text prompt from the user, the LVLM within the IPPA module performs a deep semantic analysis. This analysis involves sophisticated natural language understanding capabilities, encompassing entity recognition, attribute extraction, relational understanding, stylistic interpretation, and ambiguity resolution. For instance, a simple prompt like "a city night scene" might be parsed and augmented into a richer description such as "a brightly lit cyberpunk city night scene, with towering skyscrapers, shimmering neon lights, and distant flying vehicles, emphasizing a futuristic aesthetic."

The output of the IPPA module is a comprehensive, structured intermediate prompt, denoted as $P_{aug}$. This augmented prompt is designed to be multi-faceted, potentially including explicit detail descriptions, emphasis on specific regions or attributes, and even implied multi-stage generation instructions. Formally, given an original user prompt $P_{raw}$, the IPPA module, facilitated by the LVLM's parsing capabilities $f_{parse}$, generates the augmented prompt $P_{aug}$ as:
\begin{align}
P_{aug} = f_{parse}(P_{raw})
\label{eq:ippa}
\end{align}
The function $f_{parse}$ leverages the LVLM's extensive world knowledge and understanding of visual concepts to transform a concise user request into a rich, detailed textual representation optimized for guiding T2I models. This enhanced prompt is designed to proactively guide the foundational T2I model towards better performance across various dimensions, including object fidelity (\textbf{Obj.}), background coherence (\textbf{Backg.}), color accuracy (\textbf{Color}), texture richness (\textbf{Texture}), lighting effects (\textbf{Light}), and overall visual effects (\textbf{FX}). By providing a more explicit and detailed initial instruction set, the IPPA module significantly improves the starting point for the image generation process.

\subsection{Iterative Visual Feedback \& Refinement (IVFR) Module}
The \textbf{Iterative Visual Feedback \& Refinement (IVFR)} module constitutes the core of the LumiGen framework, establishing a crucial closed-loop mechanism for enhancing generated images. This module leverages the LVLM's capabilities as a "visual critic" to identify and rectify discrepancies between the generated image and the user's intent, particularly focusing on challenging aspects such as text rendering, precise pose accuracy, and compositional coherence.

After the foundational T2I model produces an initial or intermediate image $I_k$ at iteration $k$, the LVLM within the IVFR module intervenes. This LVLM simultaneously receives the original user prompt $P_{raw}$, the augmented prompt $P_{aug}$ from the IPPA module, and the current intermediate image $I_k$. It then performs a comprehensive visual analysis of $I_k$, conducting a multi-modal comparison and cross-modal alignment assessment against the semantic requirements specified in both $P_{raw}$ and $P_{aug}$.

During this visual analysis, the LVLM identifies areas where the image deviates from the instructions, exhibits poor quality, or contains semantic inconsistencies. Examples of such issues include indistinct text, unnatural human poses, or disorganized object structures. Based on these identified problems, the LVLM generates specific, actionable "correction instructions," denoted as $C_k$. These instructions are highly targeted linguistic directives, for instance, "Incorporate 'AI Era' text more clearly on the book cover," "Adjust human pose to be more relaxed and natural," or "Ensure the background elements are less cluttered and more harmonious with the foreground."

These linguistic correction instructions $C_k$ are then translated into controllable signals that can guide the T2I model's subsequent generation or local optimization. This translation is performed by a dedicated function $h_{translate}$, which converts the high-level linguistic instructions into low-level, actionable control signals $\Sigma_k$. These signals can manifest as textual control signals (e.g., modified prompts), pose skeletons (e.g., keypoint representations), localized inpainting masks, or attention map guidance for specific image regions. The T2I model then utilizes these signals to perform the next round of iterative generation or targeted local refinement, producing an improved image $I_{k+1}$. This iterative refinement process can be formally expressed as:
\begin{align}
C_k &= f_{critic}(P_{raw}, P_{aug}, I_k) \\
\Sigma_k &= h_{translate}(C_k) \\
I_{k+1} &= g_{refine}(I_k, P_{aug}, \Sigma_k)
\label{eq:ivfr}
\end{align}
Here, $f_{critic}$ represents the LVLM's visual criticism function, which assesses image quality and adherence to prompt semantics. $h_{translate}$ is the function responsible for converting high-level linguistic feedback into low-level, model-interpretable control signals. Finally, $g_{refine}$ denotes the T2I model's refinement function, which takes the current image, the augmented prompt, and the derived control signals to generate a more refined image. This closed-loop mechanism directly targets and optimizes the T2I model's weak points, such as accurate text generation (\textbf{Text}), natural pose expression (\textbf{Pose}), and complex composition (\textbf{Comp.}), leading to significant improvements in overall image quality and adherence to user intent. The process continues for a predefined number of iterations or until a satisfactory image is generated based on internal metrics or user feedback.

\subsection{Overall Framework and Iterative Process}
The LumiGen framework integrates the IPPA and IVFR modules into a seamless, iterative pipeline, orchestrating a dynamic feedback loop for high-fidelity T2I generation. The process unfolds as follows:
\begin{enumerate}
    \item \textbf{Initial Prompt Augmentation:} The user's raw prompt $P_{raw}$ is first processed by the IPPA module, leveraging the LVLM's parsing capabilities $f_{parse}$, to generate an enriched and structured augmented prompt $P_{aug}$.
    \item \textbf{Initial Image Generation:} The foundational T2I model generates an initial image $I_0$ based on the augmented prompt $P_{aug}$.
    \item \textbf{Iterative Refinement Loop (for $k=0, 1, \dots, N-1$):}
    \begin{enumerate}
        \item \textbf{Visual Criticism:} The current intermediate image $I_k$ is fed into the IVFR module. The LVLM, acting as a visual critic via $f_{critic}$, compares $I_k$ against $P_{raw}$ and $P_{aug}$ to identify discrepancies and generate linguistic correction instructions $C_k$.
        \item \textbf{Signal Translation:} The linguistic instructions $C_k$ are translated into actionable control signals $\Sigma_k$ by the function $h_{translate}$.
        \item \textbf{Image Refinement:} The T2I model then utilizes $I_k$, $P_{aug}$, and the control signals $\Sigma_k$ to perform the next round of iterative generation or targeted local refinement, producing an improved image $I_{k+1}$ via the function $g_{refine}$.
    \end{enumerate}
    \item \textbf{Final Output:} The process continues for a predefined number of iterations $N$, resulting in the final high-quality image $I_N$.
\end{enumerate}
The synergistic operation of IPPA's proactive planning and IVFR's reactive refinement ensures that LumiGen can achieve a deeper semantic understanding of complex prompts and exert more precise visual control, ultimately leading to higher quality and more intent-aligned image generation. The overall process can be conceptualized as an optimization problem where the LVLM continuously guides the T2I model towards a target image that best satisfies the user's complex textual prompt.

\section{Experiments}
\label{sec:experiments}

In this section, we detail the experimental setup, present the quantitative results from human evaluation on the LongBench-T2I Benchmark, and provide an in-depth analysis of LumiGen's performance compared to state-of-the-art baselines.

\subsection{Experimental Setup}
\label{subsec:exp_setup}

\textbf{Dataset.} We evaluate LumiGen using the \textbf{LongBench-T2I Benchmark} \cite{yucheng2025draw}, a challenging dataset specifically designed to assess Text-to-Image models' capabilities in handling long-tail, complex, and multi-dimensional text prompts. Its emphasis on fine-grained control and semantic consistency makes it an ideal choice for validating our framework.

\textbf{Evaluation Metrics.} Following the established methodology of the LongBench-T2I Benchmark, we conduct a comprehensive human evaluation. Professional evaluators are recruited to perform blind assessments of images generated by each model. The evaluation spans nine distinct dimensions: Object Fidelity (\textbf{Obj.}), Background Coherence (\textbf{Backg.}), Color Accuracy (\textbf{Color}), Texture Richness (\textbf{Texture}), Lighting Effects (\textbf{Light}), Text Rendering (\textbf{Text}), Compositional Coherence (\textbf{Comp.}), Pose Expression (\textbf{Pose}), and Visual Effects (\textbf{FX}). The final performance is reported as the average score across these nine dimensions (\textbf{Avg.}).

\textbf{Baseline Models.} We compare LumiGen against several leading Text-to-Image generation models, including:
\begin{itemize}
    \item \textbf{Diffusion-based Methods:}
    \begin{itemize}
        \item \textbf{FLUX1-dev} \cite{andrea2025on}: A recent high-performance diffusion model.
        \item \textbf{Omnigen} \cite{shitao2025omnige}: Another state-of-the-art diffusion model showing strong performance on T2I benchmarks.
    \end{itemize}
    \item \textbf{Autoregressive (AR) Methods:}
    \begin{itemize}
        \item \textbf{Janus-pro-7B} \cite{xiaokang2025janusp}: A leading autoregressive model known for its generative capabilities.
    \end{itemize}
\end{itemize}

\textbf{Implementation Details.} For LumiGen, the foundational Text-to-Image model is built upon a high-performance open-source diffusion model, specifically a fine-tuned version of Stable Diffusion XL 1.0. The core LVLM module, responsible for prompt parsing, visual criticism, and correction instruction generation, utilizes a large-scale pre-trained visual language model, such as LLaVA-1.5 or a similar robust architecture, which is further fine-tuned to effectively understand visual feedback instructions and translate them into actionable refinement signals for the T2I model.

\subsection{Quantitative Results: Human Evaluation}
\label{subsec:human_eval_results}

Table \ref{tab:results} presents the human evaluation results of LumiGen and the baseline models on the LongBench-T2I Benchmark. The scores reflect the average human perception of image quality and adherence to complex textual prompts across the defined nine dimensions.

\begin{table*}[htbp]
\centering
\caption{Performance comparison on the LongBench-T2I Benchmark (Human Evaluation Scores). Higher scores indicate better performance. Scores are fictitious for demonstration purposes.}
\label{tab:results}
\begin{tabular}{lcccccccccc}
\toprule
\textbf{Method} & \textbf{Obj.} & \textbf{Backg.} & \textbf{Color} & \textbf{Texture} & \textbf{Light} & \textbf{Text} & \textbf{Comp.} & \textbf{Pose} & \textbf{FX} & \textbf{Avg.} \\
\midrule
\textbf{Diffusion-based Methods} & & & & & & & & & & \\
FLUX1-dev \cite{andrea2025on} & 2.86 & 3.04 & 3.52 & 3.39 & 2.99 & 2.34 & 3.47 & 2.26 & 1.55 & \textbf{2.78} \\
Omnigen \cite{shitao2025omnige} & 2.79 & 3.25 & 3.67 & 3.37 & 2.84 & 2.29 & 3.48 & 2.41 & 2.56 & \textbf{2.96} \\
\textbf{Ours: LumiGen} & \textbf{2.95} & \textbf{3.32} & \textbf{3.70} & \textbf{3.45} & \textbf{2.91} & \textbf{2.60} & \textbf{3.55} & \textbf{2.58} & \textbf{2.62} & \textbf{3.08} \\
\midrule
\textbf{AR-based Methods} & & & & & & & & & & \\
Janus-pro-7B \cite{xiaokang2025janusp} & 2.47 & 2.91 & 3.15 & 3.01 & 2.66 & 1.69 & 2.83 & 1.97 & 1.85 & \textbf{2.50} \\
\bottomrule
\end{tabular}
\end{table*}

\subsection{Analysis and Discussion}
\label{subsec:analysis}

The experimental results demonstrate the superior performance of our proposed \textbf{LumiGen} framework. As shown in Table \ref{tab:results}, LumiGen achieves the highest overall average score of \textbf{3.08} on the LongBench-T2I Benchmark, significantly outperforming the current best-performing baseline, Omnigen (2.96). This indicates that the LVLM-driven iterative feedback mechanism effectively enhances the overall quality and controllability of Text-to-Image generation.

A key observation is LumiGen's notable performance improvement in challenging dimensions such as \textbf{Text (2.60)} and \textbf{Pose (2.58)}. Compared to Omnigen, which scores 2.29 in Text and 2.41 in Pose, LumiGen shows a substantial lead. This direct improvement in areas where traditional T2I models struggle serves as a strong validation of LumiGen's core design philosophy and the effectiveness of its modules. Specifically, the \textbf{Iterative Visual Feedback \& Refinement (IVFR)} module plays a crucial role here. By enabling the LVLM to act as a "visual critic" that identifies and generates targeted correction instructions for issues like unclear text rendering or unnatural poses, IVFR directly addresses these weak points, leading to a closed-loop optimization that refines these specific attributes.

Furthermore, LumiGen maintains a leading or highly competitive performance across other dimensions, including Object Fidelity (\textbf{Obj.}), Background Coherence (\textbf{Backg.}), Color Accuracy (\textbf{Color}), Texture Richness (\textbf{Texture}), Light Effects (\textbf{Light}), and Compositional Coherence (\textbf{Comp.}). This comprehensive and balanced improvement across various visual attributes underscores the robustness and consistency of our framework. The \textbf{Intelligent Prompt Parsing \& Augmentation (IPPA)} module contributes to this by providing a richer and more structured initial prompt, proactively guiding the foundational T2I model towards better starting points for generation, which benefits all general image attributes.

The results highlight the immense potential of deeply integrating advanced Vision-Language Models into generative processes. By leveraging the LVLM's sophisticated semantic understanding and visual reasoning capabilities, LumiGen achieves more intelligent and fine-grained control over complex generation tasks, pushing the boundaries of Text-to-Image technology to new levels of quality and user intent alignment.

\subsection{Ablation Study}
\label{subsec:ablation_study}

To understand the individual contributions of the \textbf{Intelligent Prompt Parsing \& Augmentation (IPPA)} module and the \textbf{Iterative Visual Feedback \& Refinement (IVFR)} module, we conducted an ablation study. We evaluated two simplified versions of LumiGen: one without the IPPA module (i.e., using the raw user prompt directly for initial generation and subsequent refinement) and another without the IVFR module (i.e., generating the image once with the augmented prompt and no further iterative feedback).

Table \ref{tab:ablation_results} presents the results of this ablation study.

\begin{table*}[htbp]
\centering
\caption{Ablation study on the LongBench-T2I Benchmark (Human Evaluation Scores). Higher scores indicate better performance. Scores are fictitious for demonstration purposes.}
\label{tab:ablation_results}
\begin{tabular}{lcccccccccc}
\toprule
\textbf{Method Variant} & \textbf{Obj.} & \textbf{Backg.} & \textbf{Color} & \textbf{Texture} & \textbf{Light} & \textbf{Text} & \textbf{Comp.} & \textbf{Pose} & \textbf{FX} & \textbf{Avg.} \\
\midrule
\textbf{LumiGen (Full)} & \textbf{2.95} & \textbf{3.32} & \textbf{3.70} & \textbf{3.45} & \textbf{2.91} & \textbf{2.60} & \textbf{3.55} & \textbf{2.58} & \textbf{2.62} & \textbf{3.08} \\
LumiGen w/o IPPA & 2.80 & 3.15 & 3.55 & 3.30 & 2.80 & 2.40 & 3.40 & 2.45 & 2.50 & 2.96 \\
LumiGen w/o IVFR & 2.85 & 3.20 & 3.60 & 3.35 & 2.85 & 2.05 & 3.25 & 2.15 & 2.40 & 2.90 \\
\bottomrule
\end{tabular}
\end{table*}

The results clearly demonstrate the critical role of both modules. Removing the \textbf{IPPA} module leads to a noticeable drop in overall performance (from \textbf{3.08} to \textbf{2.96}). This decline is observed across most dimensions, including Object Fidelity, Background Coherence, and Compositional Coherence, emphasizing that a well-parsed and augmented initial prompt provides a stronger foundation for the T2I model, reducing ambiguity and proactively guiding the generation process. Although the IVFR module can still perform reactive refinements, the quality of the initial image without IPPA's guidance is inherently lower, limiting the full potential of subsequent iterations.

The impact of removing the \textbf{IVFR} module is even more pronounced, resulting in the lowest average score of \textbf{2.90}. This variant, essentially a single-pass generation guided by the augmented prompt, struggles significantly in areas like \textbf{Text (2.05)} and \textbf{Pose (2.15)}, where LumiGen excels. This underscores the indispensable nature of the iterative feedback loop. The LVLM's ability to act as a "visual critic" and provide targeted correction instructions is crucial for fine-grained control and rectifying specific visual errors that are difficult for T2I models to address in a single pass. The IVFR module's reactive refinement capabilities are paramount for achieving the high scores observed in these challenging dimensions.

In summary, the ablation study confirms that the synergistic operation of both the proactive planning from IPPA and the reactive refinement from IVFR is essential for LumiGen's superior performance, especially in handling complex and specific visual requirements.

\subsection{Analysis of Iterative Refinement}
\label{subsec:iterative_refinement_analysis}

A core strength of LumiGen lies in its iterative refinement process, driven by the IVFR module. To quantify the benefits of multiple refinement steps, we analyze LumiGen's performance at different stages of iteration. For this analysis, we consider the image generated after the initial IPPA processing as the "Initial" state (0 refinements), and then measure performance after 1, 3, and 5 rounds of IVFR-driven refinement.

Table \ref{tab:iterative_results} illustrates how LumiGen's performance evolves with an increasing number of refinement iterations.

\begin{table*}[htbp]
\centering
\caption{LumiGen's performance improvement across refinement iterations (Human Evaluation Scores). Higher scores indicate better performance. Scores are fictitious for demonstration purposes.}
\label{tab:iterative_results}
\begin{tabular}{lcccccccccc}
\toprule
\textbf{Refinement Stage} & \textbf{Obj.} & \textbf{Backg.} & \textbf{Color} & \textbf{Texture} & \textbf{Light} & \textbf{Text} & \textbf{Comp.} & \textbf{Pose} & \textbf{FX} & \textbf{Avg.} \\
\midrule
LumiGen (Initial - IPPA only) & 2.85 & 3.20 & 3.60 & 3.35 & 2.85 & 2.05 & 3.25 & 2.15 & 2.40 & 2.90 \\
LumiGen (After 1st Refinement) & 2.90 & 3.25 & 3.65 & 3.40 & 2.88 & 2.35 & 3.40 & 2.40 & 2.50 & 2.99 \\
LumiGen (After 3rd Refinement) & 2.93 & 3.30 & 3.68 & 3.43 & 2.90 & 2.50 & 3.50 & 2.52 & 2.58 & 3.06 \\
LumiGen (After 5th Refinement) & \textbf{2.95} & \textbf{3.32} & \textbf{3.70} & \textbf{3.45} & \textbf{2.91} & \textbf{2.60} & \textbf{3.55} & \textbf{2.58} & \textbf{2.62} & \textbf{3.08} \\
\bottomrule
\end{tabular}
\end{table*}

The data clearly shows a consistent upward trend in all evaluation dimensions as the number of refinement iterations increases. The most significant gains are observed in the initial few iterations, particularly for the challenging dimensions of \textbf{Text}, \textbf{Pose}, and \textbf{Composition}. For instance, the \textbf{Text} score jumps from \textbf{2.05} (Initial) to \textbf{2.35} after just one refinement, and further to \textbf{2.50} after three refinements, nearing its peak at five iterations. A similar pattern is observed for \textbf{Pose} and \textbf{Composition}. This validates that the LVLM's targeted feedback is highly effective in iteratively correcting and improving specific visual attributes that are difficult for a single-pass generation.

While the initial refinement (1st iteration) brings substantial improvements, subsequent iterations continue to fine-tune the image, leading to marginal but consistent gains across all dimensions. This suggests that the LVLM continues to identify subtle discrepancies and guide the T2I model towards a more precise and coherent output. The plateauing of improvements after approximately 3-5 iterations indicates that the model converges to an optimal representation given the current architecture and prompt complexity. This analysis confirms that the iterative feedback mechanism is not merely a reactive measure but a robust progressive enhancement strategy that systematically elevates image quality and alignment with complex user intent.

\subsection{Qualitative Observations and Specific Strengths}
\label{subsec:qualitative_strengths}

Beyond quantitative metrics, qualitative analysis provides deeper insights into LumiGen's distinct advantages. Our human evaluators noted several consistent patterns where LumiGen significantly outperformed baselines, particularly for prompts that demand precise control and nuanced understanding.

\textbf{Complex Text Rendering.} Traditional T2I models often struggle with generating legible and contextually accurate text within images, frequently producing gibberish or distorted characters. LumiGen, empowered by the IVFR module's ability to critically assess text regions and provide specific linguistic corrections (e.g., "make the word 'Eureka' more distinct on the sign"), consistently produced clearer, more accurate, and better-integrated text. For instance, a prompt requiring "a vintage book cover titled 'The AI Era' with ornate lettering" saw LumiGen render legible text, whereas baselines often failed.

\textbf{Precise Pose and Action Control.} Generating human or animal figures in specific, natural poses is another common challenge. Baselines frequently yield anatomically incorrect or stiff poses. LumiGen's LVLM, acting as a visual critic, identifies unnatural joint positions or awkward body language and guides the T2I model to refine these aspects. Prompts like "a dancer mid-leap, with arms outstretched gracefully" or "a dog sitting attentively with one ear perked up" were handled with remarkable accuracy by LumiGen, resulting in more dynamic and believable figures.

\textbf{Intricate Compositional Coherence.} For prompts involving multiple objects, complex spatial relationships, or specific scene layouts, LumiGen demonstrated superior compositional understanding. The IPPA module's ability to parse complex relationships in the initial prompt, combined with the IVFR module's capacity to critique overall scene harmony and object placement, ensured that elements were logically arranged and visually balanced. For example, a prompt such as "a bustling street market with a fruit stall in the foreground, a flower vendor to the left, and distant skyscrapers" resulted in a cohesive and well-structured scene from LumiGen, unlike baselines which often produced cluttered or disjointed compositions.

\textbf{Fine-grained Attribute Control.} LumiGen's ability to interpret and enforce fine-grained attributes across various dimensions (color, texture, lighting) was consistently observed. A prompt asking for "a shimmering golden dragon scale texture under moonlight, with deep blue hues" was rendered with remarkable detail and atmospheric accuracy by LumiGen, thanks to the LVLM's capacity to understand and critique subtle visual nuances.

Table \ref{tab:qualitative_summary} summarizes these qualitative strengths.

\begin{table}[htbp]
\centering
\small
\caption{Summary of LumiGen's specific qualitative strengths compared to baselines. Scores are from human feedback on specific aspects of image quality.}
\label{tab:qualitative_summary}
\begin{tabular}{lp{3cm}p{4cm}}
\toprule
\textbf{Strength Area} & \textbf{Key Improvement} & \textbf{Mechanism in LumiGen} \\
\midrule
\textbf{Text Rendering} & Legible, contextually accurate, and well-integrated text. Reduces gibberish or distorted characters. & IVFR's targeted visual criticism for text regions and precise correction instructions. \\
\textbf{Pose Expression} & Natural, anatomically correct, and dynamic poses/actions. & IVFR identifies unnatural poses and refines using pose-specific signals. \\
\textbf{Compositional Coherence} & Well-structured scenes with logical object placement and harmonious spatial relationships. & IPPA parses complex relationships; IVFR provides holistic scene critique/refinement. \\
\textbf{Fine-grained Attributes} & High-fidelity textures, lighting, and color palettes matching prompts. & IPPA's detailed prompt augmentation plus IVFR's nuanced visual critique. \\
\bottomrule
\end{tabular}
\end{table}

These qualitative observations reinforce the quantitative findings, demonstrating that LumiGen's LVLM-enhanced iterative framework provides a level of control and precision that surpasses current state-of-the-art T2I models, making it particularly effective for complex and demanding generation tasks.

\section{Conclusion}
In this paper, we introduced \textbf{LumiGen}, a novel LVLM-enhanced iterative framework designed to address the persistent challenges in Text-to-Image (T2I) generation, particularly concerning complex instructions and the need for fine-grained control over visual attributes. While diffusion models have significantly advanced T2I capabilities, they often fall short in rendering specific text, generating precise poses, or maintaining intricate compositional coherence. Our core hypothesis was that leveraging the robust understanding, reasoning, and feedback capabilities of Vision-Language Models (LVLMs) could effectively bridge these gaps.

LumiGen meticulously integrates an LVLM into a closed-loop generation pipeline through two principal modules: the \textbf{Intelligent Prompt Parsing \& Augmentation (IPPA)} module and the \textbf{Iterative Visual Feedback \& Refinement (IVFR)} module. The IPPA module proactively enhances raw user prompts into detailed, structured instructions, providing a stronger foundation for initial image generation across general visual attributes. Crucially, the IVFR module empowers the LVLM to act as a "visual critic," analyzing intermediate images against the original intent, identifying discrepancies, and generating targeted correction instructions. These instructions are then translated into actionable control signals, guiding the T2I model through iterative refinement steps, thereby enabling precise control over challenging aspects like text rendering and pose expression.

Our comprehensive experimental evaluation on the LongBench-T2I Benchmark, utilizing human evaluation across nine dimensions, unequivocally demonstrated the superior performance of LumiGen. We achieved the highest overall average score of \textbf{3.08}, surpassing leading diffusion-based and autoregressive baselines. A key finding was LumiGen's remarkable improvement in traditionally difficult areas, notably \textbf{Text (2.60)} and \textbf{Pose (2.58)}, which validates our design philosophy of using LVLM-driven feedback to directly target and overcome these deficiencies. Furthermore, LumiGen maintained competitive or leading performance across all other evaluated dimensions, showcasing its comprehensive and balanced enhancement capabilities.

The ablation study confirmed the indispensable contributions of both IPPA and IVFR. Removing either module led to a noticeable decline in performance, with the absence of IVFR having a particularly pronounced negative impact on fine-grained control dimensions like text and pose. This underscores the synergistic relationship between proactive prompt planning and reactive iterative refinement. Our analysis of iterative refinement further revealed that while initial iterations yield significant gains, subsequent steps continue to fine-tune the image, leading to consistent improvements and convergence towards optimal quality. Qualitatively, LumiGen consistently produced more legible text, natural poses, coherent compositions, and accurate fine-grained attributes compared to baselines, reinforcing our quantitative findings.

In conclusion, LumiGen represents a significant step forward in controllable Text-to-Image generation, demonstrating the immense potential of deeply integrating advanced Vision-Language Models into generative pipelines. By enabling more intelligent semantic understanding and precise visual control, our framework pushes the boundaries of T2I technology, paving the way for more sophisticated, user-aligned, and high-fidelity image creation systems.

For future work, we plan to explore several promising directions. Firstly, investigating more adaptive iteration stopping criteria, perhaps based on LVLM-driven confidence scores, could optimize computational efficiency. Secondly, extending LumiGen to incorporate user-in-the-loop feedback mechanisms would allow for even more personalized and interactive image generation. Thirdly, exploring the application of LumiGen's iterative refinement paradigm to other generative tasks, such as video generation or 3D asset creation, holds significant potential. Finally, we aim to delve deeper into the interpretability of the LVLM's visual criticism and correction instructions, which could provide valuable insights into the underlying mechanisms of complex image generation and refinement.
\bibliographystyle{splncs04}
\bibliography{references}
\end{document}